\documentclass[12pt,american,english]{article}

\usepackage[latin9]{inputenc}
\usepackage{geometry}
\usepackage{amsmath}
\usepackage{amssymb}
\usepackage{graphicx}
\usepackage{setspace}
\makeatletter
\@ifundefined{date}{}{\date{}}
\usepackage{array}
\usepackage{mathrsfs}
\usepackage{color}

\makeatother

\usepackage{babel}
\begin{document}

\title{\textbf{A Criterion for Extending}\\
\textbf{Continuous-Mixture Identifiability Results}}

\author{Michael R. Powers\thanks{Corresponding author; Department of Finance, School of Economics and
Management, and Schwarzman College, Tsinghua University, Beijing,
China 100084; email: powers@sem.tsinghua.edu.cn.} \ and Jiaxin Xu\thanks{Organization Department, CPC Yichang Municipal Committee, Yichang,
Hubei, China; email: jiaxinxucq@163.com.}}

\date{June 16, 2025}
\maketitle
\begin{abstract}
\begin{singlespace}
\noindent Mixture distributions provide a versatile and widely used
framework for modeling random phenomena, and are particularly well-suited
to the analysis of geoscientific processes and their attendant risks
to society. For continuous mixtures of random variables, we specify
a simple criterion \textendash{} \emph{generating-function accessibility}
\textendash{} to extend previously known kernel-based identifiability
(or unidentifiability) results to new kernel distributions. This criterion,
based on functional relationships between the relevant kernels' moment-generating
functions or Laplace transforms, may be applied to continuous mixtures
of both discrete and continuous random variables. To illustrate the
proposed approach, we present results for several specific kernels,
in each case briefly noting its relevance to research in the geosciences
and/or related risk analysis.\medskip{}

\end{singlespace}
\end{abstract}

\section{Introduction}

\noindent Mixture distributions provide a versatile and widely used
framework for modeling random phenomena, characterized by both: (i)
a theoretical structure capable of incorporating an array of coexisting
processes differentiated by random perturbations; and (ii) the practical
flexibility to replicate empirically observed patterns through a set
of adjustable components. These characteristics make such models especially
well suited for the highly complex and chaotic physical processes
of the geosciences (e.g., seismology, hydrology, and meteorology),
as well as the risks to society posed by these processes (e.g., earthquakes,
floods, and cyclones).

Of particular note is the stream of research initiated by Vere-Jones
and Davies (1966) and Shlien and Toksöz (1970), who applied the ``cluster
model'' of Neyman and Scott (1952) to earthquake arrival times through
mixtures of Poisson processes. Over the past five decades, cluster
models have been employed by researchers in many geoscientific fields,
including hydrology (e.g., Cowpertwait and O\textquoteright Connell,
1997), climatology (e.g., Dacre and Pinto, 2020), and meteorology
(e.g., Chaturvedi, et al., 2025). At the same time, other researchers,
such as Bhatti (2004), Tanaka, Ogata, and Stoyan (2008), and Wang
et al. (2024), have developed new and increasingly sophisticated statistical
methods to analyze data within the cluster-model setting. 

In the field of risk analysis, researchers typically study random
processes resulting in potential financial or other damage to society
by disaggregating them\foreignlanguage{american}{ into two distinct
components: \emph{counts}, the unknown numbers of damage-causing events,
modeled as nonnegative discrete random variables; and \emph{magnitudes},
the unknown damage amounts arising from the respective events, modeled
as nonnegative continuous random variables.}\footnote{\selectlanguage{american}%
The terms ``counts'' and ``magnitudes'' often are replaced by
more specialized terms in certain disciplines. In particular, insurance
researchers invariably refer to counts as \emph{frequencies} and to
magnitudes as \emph{severities}, whereas a variety of less established
terms are used in other fields (such as \emph{rate}, \emph{probability},
or \emph{likelihood} for count, and \emph{intensity}, \emph{impact},
or \emph{consequence} for magnitude).\selectlanguage{english}%
}\foreignlanguage{american}{ Both components are readily modeled by
mixture distributions to capture heterogeneity within their respective
risk processes, and can be treated within the same theoretical scheme
(see }Powers and Xu, 2025)\foreignlanguage{american}{. Examples of
mixture-distribution applications involving risks generated by geophysical
processes include the work of }Mak, Bingham, and Lu (2016) and Li,
Tang, and Jiang (2016), whereas broad-based methodological analyses
have been given by Younis, Aslam, and Bhatti (2021) and Deng and Aminzadeh
(2023).

Regardless of the particular application involved, it usually is desirable
(if not crucial) to be able to associate any given mixture distribution
with a unique mixing distribution (and vice versa). This property
is known as \emph{identifiability}. The problem of mixture identifiability
enjoys a long and interesting research literature, but is somewhat
limited in terms of results with broad applicability. In most cases,
researchers focus on one of two general categories of mixtures \textendash{}
those with discrete mixing distributions or continuous mixing distributions
\textendash{} and then seek sufficient conditions for either the kernel
distribution or the mixing distribution to generate identifiable mixtures.
Karlis and Xekalaki (2003) discuss the importance of identifiability
in mixture models and Maritz and Lwin (1989), Prakasa Rao (1992),
and Stoyanov and Lin (2011) provide general overviews of the relevant
literature. More detailed references are given in Section 2 below.

The present article addresses continuous-mixture identifiability (and
unidentifiability) results based on properties of the kernel distribution.
We begin by summarizing the currently known sufficient conditions
for identifiability in Section 2, and then provide a simple criterion
to extend previously known results to new kernel distributions in
Section 3. This criterion, called \emph{generating-function (GF) accessibility},
is based on functional relationships between the relevant kernels'
moment-generating functions (MGFs) or Laplace transforms, and may
be applied to continuous mixtures of both discrete and continuous
random variables. We illustrate the proposed approach in Section 4
by presenting results that demonstrate the identifiability or unidentifiability
of continuous mixtures generated by several specific kernel distributions.
In each case, the kernel's relevance to research in the geosciences
and/or related risk analysis is briefly noted.

\section{Identifiable Continuous Mixtures}

\noindent Let $X\mid\boldsymbol{\alpha};\boldsymbol{\alpha^{*}}\sim F_{X\mid\boldsymbol{\alpha};\boldsymbol{\alpha^{*}}}\left(x\right),\:x\in\mathfrak{X}\subseteq\mathbb{R}$,
where $F_{X\mid\boldsymbol{\alpha};\boldsymbol{\alpha^{*}}}$ is a
discrete or continuous cumulative distribution function (CDF) with
\emph{free parameter} vector $\boldsymbol{\alpha}=\left(\alpha_{1},\ldots,\alpha_{k}\right)$
(to be treated as random for mixing purposes) and \emph{fixed parameter}
vector $\boldsymbol{\alpha^{*}}=\left(\alpha_{1}^{*},\ldots,\alpha_{\ell}^{*}\right)$.
For clarity, we will write $\alpha_{i}\in\mathcal{A}_{i}\subseteq\mathbb{R}$
for $i\in\left\{ 1,\ldots,k\right\} $ and $\alpha_{j}^{*}\in\mathcal{A}_{j}^{*}\subseteq\mathbb{R}$
for $j\in\left\{ 1,\ldots,\ell\right\} $.

Now consider a generic continuous-mixture model,
\begin{equation}
F_{X\mid\boldsymbol{\alpha^{*}}}\left(x\right)={\textstyle {\displaystyle \int_{\mathcal{\boldsymbol{A}}\left(k\right)}}}F_{X\mid\boldsymbol{\alpha};\boldsymbol{\alpha^{*}}}\left(x\right)g_{\boldsymbol{\alpha}}\left(\boldsymbol{\alpha}\right)d\boldsymbol{\alpha},
\end{equation}
where $g_{\boldsymbol{\alpha}}$ is a joint probability density function
(PDF) for $\boldsymbol{\alpha}\in\boldsymbol{\mathcal{A}}\left(k\right)=\mathcal{A}_{1}\times\cdots\times\mathcal{A}_{k}$.
Using conventional terminology, we will call $F_{X\mid\boldsymbol{\alpha};\boldsymbol{\alpha^{*}}}$
the \emph{kernel} CDF, $g_{\boldsymbol{\alpha}}$ the \emph{mixing}
PDF, and $F_{X\mid\boldsymbol{\alpha^{*}}}$ the \emph{mixture} (or
\emph{mixed}) CDF.

As noted in the Introduction, the property of mixture identifiability
\textendash{} that is, the existence of a unique mixing distribution
(represented here by $g_{\boldsymbol{\alpha}}$)\footnote{\noindent We will say that $g_{\boldsymbol{\alpha}}$ is \emph{not
unique} if there exist at least two PDFs, $g_{\boldsymbol{\alpha}}^{\left(\textrm{I}\right)}\left(\boldsymbol{\alpha}\right)\textrm{ and }g_{\boldsymbol{\alpha}}^{\left(\textrm{II}\right)}\left(\boldsymbol{\alpha}\right),\:\boldsymbol{\alpha}\in\boldsymbol{\mathcal{A}}\left(k\right)$,
such that both may be substituted for $g_{\boldsymbol{\alpha}}\left(\boldsymbol{\alpha}\right)$
in (1), and $g_{\boldsymbol{\alpha}}^{\left(\textrm{I}\right)}\left(\boldsymbol{\alpha}\right)\neq g_{\boldsymbol{\alpha}}^{\left(\textrm{II}\right)}\left(\boldsymbol{\alpha}\right)$
for all $\boldsymbol{\alpha}$ within some continuous region $\boldsymbol{\mathcal{A}}\subseteq\boldsymbol{\mathcal{A}}\left(k\right)$
of dimension $k$.} for a given mixture distribution (represented by $F_{X\mid\boldsymbol{\alpha^{*}}}$)
\textendash{} is central to modeling applications. As also noted,
our objective is to provide a criterion for extending previously known
continuous-mixture identifiability (and unidentifiability) results
based on properties of the kernel distribution. In particular, we
envision applying the criterion to kernels in one (or more) of the
following major categories for which continuous-mixture identifiability
has been established.

\subsection{Poisson Distribution}

\noindent If $X\mid\lambda\sim\textrm{Poisson}\left(\lambda\right)$
with free parameter $\lambda\in\mathbb{R}^{+}$, then
\[
F_{X}\left(x\right)={\textstyle {\displaystyle \int_{0}^{\infty}}}F_{X\mid\lambda}^{\left(\textrm{P}\right)}\left(x\right)g_{\lambda}\left(\lambda\right)d\lambda
\]
is identifiable. This classic result of Feller (1943) is of fundamental
importance because of the broad application of Poisson kernels in
mixture models.

\subsection{Additively Closed Family}

\noindent Let $X\mid r;\boldsymbol{\alpha^{*}}\sim F_{X\mid r;\boldsymbol{\alpha^{*}}}^{\left(\textrm{AC}\right)}\left(x\right),\:x\in\mathfrak{X}\subseteq\mathbb{R}$
be a discrete or continuous random variable with free parameter $r\in\mathbb{R}^{+}$
and fixed parameter vector $\boldsymbol{\alpha^{*}}$, where the associated
characteristic function (CF) may be written as $\varphi_{X\mid r;\boldsymbol{\alpha^{*}}}\left(\omega\right)=\textrm{E}_{X\mid r;\boldsymbol{\alpha^{*}}}\left[e^{i\omega X}\right]=\left[\varrho\left(\omega,\boldsymbol{\alpha^{*}}\right)\right]^{r}$,
for some complex-valued function $\varrho$. Then $X$ belongs to
the additively closed family, and
\[
F_{X\mid\boldsymbol{\alpha^{*}}}\left(x\right)={\textstyle {\displaystyle \int_{0}^{\infty}}}F_{X\mid r;\boldsymbol{\alpha^{*}}}^{\left(\textrm{AC}\right)}\left(x\right)g_{r}\left(r\right)dr
\]
is identifiable. (See Teicher, 1961.)

Notable special cases of the additively closed family include:
\begin{itemize}
\item $X\mid r\sim\textrm{Poisson}\left(r\right)\Longleftrightarrow F_{X\mid r}^{\left(\textrm{P}\right)}\left(x\right)={\textstyle \sum_{n=0}^{x}\tfrac{e^{-r}r^{n}}{n!}},\:x\in\mathbb{Z}_{0}^{+}$,
with free $r$;\footnote{To avoid extensive use of floor-function notation ($\left\lfloor x\right\rfloor $),
we will state functional values of discrete CDFs only at points $x$
with positive point mass (i.e., $\Pr\left\{ X=x\right\} >0$).}
\item $X\mid r;p\sim\textrm{Negative Binomial}\left(r,p\right)\Longleftrightarrow F_{X\mid r;p}^{\left(\textrm{NB}\right)}\left(x\right)={\textstyle \sum_{n=0}^{x}\tfrac{\Gamma\left(r+n\right)}{\Gamma\left(r\right)\Gamma\left(n+1\right)}\left(1-p\right)^{r}p^{n}},\:x\in\mathbb{Z}_{0}^{+}$,
with free $r$ and fixed $p\in\left(0,1\right)$; and
\item $X\mid r;\theta\sim\textrm{Gamma}\left(r,\theta\right)\Longleftrightarrow F_{X\mid r;\theta}^{\left(\Gamma\right)}\left(x\right)={\textstyle \int_{0}^{x}}\tfrac{z^{r-1}e^{-z/\theta}}{\Gamma\left(r\right)\theta^{r}}dz,\:x\in\mathbb{R}^{+}$,
with free $r$ and fixed $\theta\in\mathbb{R}^{+}$.
\end{itemize}

\subsection{Scale-Parameter Family}

\noindent Let $X\mid\theta;\boldsymbol{\alpha^{*}}\sim F_{X\mid\theta;\boldsymbol{\alpha^{*}}}^{\left(\textrm{SP}\right)}\left(x\right),\:x\in\mathfrak{X}\subseteq\mathbb{R}^{+}$
be a continuous random variable with free parameter $\theta\in\mathbb{R}^{+}$
and fixed parameter vector $\boldsymbol{\alpha^{*}}$, where $F_{X\mid\theta;\boldsymbol{\alpha^{*}}}^{\left(\textrm{SP}\right)}\left(x\right)\equiv F_{X\mid1;\boldsymbol{\alpha^{*}}}^{\left(\textrm{SP}\right)}\left(\tfrac{x}{\theta}\right)$.
Then $X$ belongs to the scale-parameter family, and
\[
F_{X\mid\boldsymbol{\alpha^{*}}}\left(x\right)={\textstyle {\displaystyle \int_{0}^{\infty}}}F_{X\mid\theta;\boldsymbol{\alpha^{*}}}^{\left(\textrm{SP}\right)}\left(x\right)g_{\theta}\left(\theta\right)d\theta
\]
is identifiable. (See Teicher, 1961.)

Notable special cases of this family include:
\begin{itemize}
\item $X\mid\theta;r\sim\textrm{Gamma}\left(r,\theta\right)\Longleftrightarrow F_{X\mid\theta;r}^{\left(\Gamma\right)}\left(x\right)={\textstyle \int_{0}^{x}}\tfrac{z^{r-1}e^{-z/\theta}}{\Gamma\left(r\right)\theta^{r}}dz,\:x\in\mathbb{R}^{+}$,
with free $\theta$ and fixed $r\in\mathbb{R}^{+}$;
\item $X\mid\theta;\tau\sim\textrm{Weibull}\left(\theta,\tau\right)\Longleftrightarrow F_{X\mid\theta;\tau}^{\left(\textrm{W}\right)}\left(x\right)=1-\exp\left(-\left(\tfrac{x}{\theta}\right)^{\tau}\right),\:x\in\mathbb{R}^{+}$,
with free $\theta$ and fixed $\tau\in\mathbb{R}^{+}$; and
\item $X\mid\theta;\alpha\sim\textrm{Pareto 1}\left(\alpha,\theta\right)\Longleftrightarrow F_{X\mid\theta;\alpha}^{\left(\textrm{P}1\right)}\left(x\right)=1-\left(\tfrac{\theta}{x}\right)^{\alpha},\:x\in\left(\theta,\infty\right)$,
with free $\theta$ and fixed $\alpha\in\mathbb{R}^{+}$.
\end{itemize}

\subsection{Location-Parameter Family}

\noindent Let $X\mid m;\boldsymbol{\alpha^{*}}\sim F_{X\mid m;\boldsymbol{\alpha^{*}}}^{\left(\textrm{LP}\right)}\left(x\right),\:x\in\mathfrak{X}\subseteq\mathbb{R}$
be a continuous random variable with free parameter $m\in\mathbb{R}$
and fixed parameter vector $\boldsymbol{\alpha^{*}}$, where $F_{X\mid m;\boldsymbol{\alpha^{*}}}^{\left(\textrm{LP}\right)}\left(x\right)\equiv F_{X\mid0;\boldsymbol{\alpha^{*}}}^{\left(\textrm{LP}\right)}\left(x-m\right)$.
Then $X$ belongs to the location-parameter family, and
\[
F_{X\mid\boldsymbol{\alpha^{*}}}\left(x\right)={\textstyle {\displaystyle \int_{-\infty}^{\infty}}}F_{X\mid m;\boldsymbol{\alpha^{*}}}^{\left(\textrm{LP}\right)}\left(x\right)g_{m}\left(m\right)dm
\]
is identifiable. (See Teicher, 1961.) One can extend this result to
the location-scale-parameter family (such that $X\mid m;\boldsymbol{\alpha^{*}}\sim F_{X\mid m;\boldsymbol{\alpha^{*}}}^{\left(\textrm{LSP}\right)}\left(x\right)\equiv F_{X\mid0;1,\alpha_{2}^{*},\ldots,\alpha_{\ell}^{*}}^{\left(\textrm{LSP}\right)}\left(\tfrac{x-m}{\varsigma}\right)$)
by introducing a fixed scale parameter, $\varsigma=\alpha_{1}^{*}\in\mathbb{R}^{+}$.

Notable members of this family include:
\begin{itemize}
\item $X\mid m;\varsigma^{2}\sim\textrm{Normal}\left(m,\varsigma^{2}\right)\Longleftrightarrow F_{X\mid m;\varsigma^{2}}^{\left(\textrm{N}\right)}\left(x\right)={\textstyle \int_{-\infty}^{x}}\tfrac{1}{\sqrt{2\pi}\varsigma}\exp\left(-\tfrac{\left(z-m\right)^{2}}{2\varsigma^{2}}\right)dz,\:x\in\mathbb{R}$,
with free $m$ and fixed $\varsigma^{2}$;
\item $X\mid m;\varsigma\sim\textrm{Laplace}\left(m,\varsigma\right)\Longleftrightarrow F_{X\mid m;\varsigma}^{\left(\textrm{L}\right)}\left(x\right)={\textstyle \int_{-\infty}^{x}}\tfrac{1}{2\varsigma}\exp\left(-\tfrac{\left|z-m\right|}{\varsigma}\right)dz,\:x\in\mathbb{R}$,
with free $m$ and fixed $\varsigma$; and
\item $X\mid m;\varsigma\sim\textrm{Gumbel}\left(m,\varsigma\right)\Longleftrightarrow F_{X\mid m;\varsigma}^{\left(\textrm{G}\right)}\left(x\right)=\exp\left(-\exp\left(-\left(\tfrac{x-m}{\varsigma}\right)\right)\right),\:x\in\mathbb{R}$,
with free $m$ and fixed $\varsigma$.
\end{itemize}

\subsection{Infinite Power-Series Family}

\noindent Let $X\mid q;\boldsymbol{\alpha^{*}}\sim F_{X\mid q;\boldsymbol{\alpha^{*}}}^{\left(\textrm{IPS}\right)}\left(x\right),\:x\in\mathfrak{X}\subseteq\mathbb{Z}_{0}^{+}$
be a discrete random variable with free parameter $q\in\mathcal{Q}\subseteq\mathbb{R}^{+}$
and fixed parameter vector $\boldsymbol{\alpha^{*}}$, where the probability
mass function (PMF) may be written as $f_{X\mid q;\boldsymbol{\alpha^{*}}}\left(x\right)=\tfrac{c_{x}q^{x}}{C\left(q\right)}$
for a sequence of real-valued constants, $c_{x}\geq0$, and $C\left(q\right)={\textstyle \sum_{x=0}^{\infty}}c_{x}q^{x}$.
Then $X$ belongs to the infinite power-series family, and 
\[
F_{X\mid\boldsymbol{\alpha^{*}}}\left(x\right)={\textstyle {\displaystyle \int_{0}^{\infty}}}F_{X\mid q;\boldsymbol{\alpha^{*}}}^{\left(\textrm{IPS}\right)}\left(x\right)g_{q}\left(q\right)dq
\]
is identifiable under certain conditions. (See Lüxmann-Ellinghaus,
1987, Sapatinas, 1995, and Stoyanov and Lin, 2011 for results with
increasingly broad conditions.)

Special cases of the infinite power-series family include:
\begin{itemize}
\item $X\mid q\sim\textrm{Poisson}\left(q\right)\Longleftrightarrow F_{X\mid q}^{\left(\textrm{P}\right)}\left(x\right)={\textstyle \sum_{n=0}^{x}\tfrac{e^{-q}q^{n}}{n!}},\:x\in\mathbb{Z}_{0}^{+}$,
with free $q\in\mathbb{R}^{+}$;
\item $X\mid q;r\sim\textrm{Negative Binomial}\left(r,q\right)\Longleftrightarrow F_{X\mid q;r}^{\left(\textrm{NB}\right)}\left(x\right)={\textstyle \sum_{n=0}^{x}\tfrac{\Gamma\left(r+n\right)}{\Gamma\left(r\right)\Gamma\left(n+1\right)}\left(1-q\right)^{r}q^{n}},\:x\in\mathbb{Z}_{0}^{+}$,
with free $q\in\left(0,1\right)$ and fixed $r\in\mathbb{R}^{+}$;
and
\item $X\mid q\sim\textrm{Logarithmic}\left(q\right)\Longleftrightarrow F_{X\mid q}^{\left(\textrm{Log}\right)}\left(x\right)=\sum_{n=1}^{x}\tfrac{-q^{n}}{n\ln\left(1-q\right)},\:x\in\mathbb{Z}^{+}$,
with free $q\in\left(0,1\right)$.
\end{itemize}

\section{Generating-Function Accessibility}

\subsection{Definition}

\noindent In conjunction with $X\mid\boldsymbol{\alpha};\boldsymbol{\alpha^{*}}\sim F_{X\mid\boldsymbol{\alpha};\boldsymbol{\alpha^{*}}}\left(x\right),\:x\in\mathfrak{X}\subseteq\mathbb{R}$,
we now consider $Y\mid\boldsymbol{\beta};\boldsymbol{\beta^{*}}\sim F_{Y\mid\boldsymbol{\beta};\boldsymbol{\beta^{*}}}\left(y\right),\:y\in\mathfrak{Y}\subseteq\mathbb{R}$,
where $F_{Y\mid\boldsymbol{\beta};\boldsymbol{\beta^{*}}}$ is a discrete
or continuous CDF with free parameter vector $\boldsymbol{\beta}=\left(\beta_{1},\ldots,\beta_{k}\right)$
and fixed parameter vector $\boldsymbol{\beta^{*}}=\left(\beta_{1}^{*},\ldots,\beta_{\nu}^{*}\right)$
(such that $\beta_{i}\in\mathcal{B}_{i}\subseteq\mathbb{R}$ for $i\in\left\{ 1,\ldots,k\right\} $
and $\beta_{j}^{*}\in\mathcal{B}_{j}^{*}\subseteq\mathbb{R}$ for
$j\in\left\{ 1,\ldots,\nu\right\} $). Furthermore, let $\textrm{M}_{X\mid\boldsymbol{\alpha};\boldsymbol{\alpha^{*}}}\left(s\right)=\textrm{E}_{X\mid\boldsymbol{\alpha};\boldsymbol{\alpha^{*}}}\left[e^{sX}\right]$
and $\textrm{M}_{Y\mid\boldsymbol{\beta};\boldsymbol{\beta^{*}}}\left(t\right)=\textrm{E}_{Y\mid\boldsymbol{\beta};\boldsymbol{\beta^{*}}}\left[e^{tY}\right]$
denote the MGFs associated with $F_{X\mid\boldsymbol{\alpha};\boldsymbol{\alpha^{*}}}$
and $F_{Y\mid\boldsymbol{\beta};\boldsymbol{\beta^{*}}}$, respectively.
This provides context for the following definition.\medskip{}

\noindent \textbf{Definition 1:} The function $\textrm{M}_{Y\mid\boldsymbol{\beta};\boldsymbol{\beta^{*}}}\left(t\right)$
is said to be \emph{GF accessible} from $\textrm{M}_{X\mid\boldsymbol{\alpha};\boldsymbol{\alpha^{*}}}\left(s\right)$
if:

(i) there exists a continuous one-to-one vector mapping, $\boldsymbol{\beta}=\boldsymbol{\eta}\left(\boldsymbol{\alpha}\right):\boldsymbol{\mathcal{A}}\left(k\right)\rightarrow\boldsymbol{\mathcal{B}}\left(k\right)$,
that may depend on $\boldsymbol{\alpha^{*}}$ and $\boldsymbol{\beta^{*}}$
but not on $s$ and $t$;\footnote{\noindent The symbol $\boldsymbol{\mathcal{B}}\left(k\right)$ is
defined analogously to $\boldsymbol{\mathcal{A}}\left(k\right)$;
i.e., $\boldsymbol{\mathcal{B}}\left(k\right)=\mathcal{B}_{1}\times\cdots\times\mathcal{B}_{k}$.}

(ii) there exists a continuous one-to-one mapping, $t=\xi\left(s\right):\left[0,\epsilon_{1}\right)\rightarrow\left[0,\epsilon_{2}\right)$
(for some fixed real values $\epsilon_{1}>0$ and $\epsilon_{2}>0$),
that may depend on $\boldsymbol{\alpha^{*}}$ and $\boldsymbol{\beta^{*}}$
but not on $\boldsymbol{\alpha}$ and $\boldsymbol{\beta}$; and

(iii) $\textrm{M}_{Y\mid\boldsymbol{\eta}\left(\boldsymbol{\alpha}\right);\boldsymbol{\beta^{*}}}\left(\xi\left(s\right)\right)=\textrm{M}_{X\mid\boldsymbol{\alpha};\boldsymbol{\alpha^{*}}}\left(s\right)$.\medskip{}

Clearly, GF accessibility is a symmetric property; that is, $\textrm{M}_{Y\mid\boldsymbol{\beta};\boldsymbol{\beta^{*}}}\left(t\right)$
is accessible from\linebreak{}
$\textrm{M}_{X\mid\boldsymbol{\alpha};\boldsymbol{\alpha^{*}}}\left(s\right)$
if and only if $\textrm{M}_{X\mid\boldsymbol{\alpha};\boldsymbol{\alpha^{*}}}\left(s\right)$
is accessible from $\textrm{M}_{Y\mid\boldsymbol{\beta};\boldsymbol{\beta^{*}}}\left(t\right)$.
Moreover, it is important to recognize that the analytical classification
(i.e., discrete or continuous) of $F_{X\mid\boldsymbol{\alpha};\boldsymbol{\alpha^{*}}}$
need not be the same as that of $F_{Y\mid\boldsymbol{\beta};\boldsymbol{\beta^{*}}}$
for GF accessibility to hold. Finally, we would note that the accessibility
concept applies equally well to Laplace transforms, which may be necessary
for cases in which the MGF is undefined because of a heavy right tail.
The extension to CFs is less straightforward because of difficulties
in finding an appropriate function $\xi$ to satisfy condition (ii)
of Definition 1.

\subsection{Main Result}

\noindent The following result justifies employing GF accessibility
as a criterion for extending an established identifiability/unidentifiability
result for one kernel, $F_{X\mid\boldsymbol{\alpha};\boldsymbol{\alpha^{*}}}$,
to the equivalent result for another kernel, $F_{Y\mid\boldsymbol{\beta};\boldsymbol{\beta^{*}}}$.

\medskip{}

\noindent \textbf{Theorem 1:} If $\textrm{M}_{Y\mid\boldsymbol{\beta};\boldsymbol{\beta^{*}}}\left(t\right)$
is GF accessible from $\textrm{M}_{X\mid\boldsymbol{\alpha};\boldsymbol{\alpha^{*}}}\left(s\right)$,
then all continuous mixtures generated by the kernel $F_{Y\mid\boldsymbol{\beta};\boldsymbol{\beta^{*}}}$
are identifiable if and only if all continuous mixtures generated
by the kernel $F_{X\mid\boldsymbol{\alpha};\boldsymbol{\alpha^{*}}}$
are identifiable.\medskip{}

\noindent \textbf{Proof:} We begin with the ``if'' portion of the
theorem (i.e., that if $F_{X\mid\boldsymbol{\alpha};\boldsymbol{\alpha^{*}}}$
generates identifiable continuous mixtures, then $F_{Y\mid\boldsymbol{\beta};\boldsymbol{\beta^{*}}}$
must do so as well), and proceed by contradiction. That is, for some
mixed CDF $F_{Y\mid\boldsymbol{\beta^{*}}}$, we assume there exist
PDFs $g_{\boldsymbol{\beta}}^{\left(\textrm{I}\right)}$ and $g_{\boldsymbol{\beta}}^{\left(\textrm{II}\right)}$
such that
\[
F_{Y\mid\boldsymbol{\beta^{*}}}\left(y\right)={\displaystyle \int_{\boldsymbol{\mathcal{B}}\left(k\right)}}F_{Y\mid\boldsymbol{\beta};\boldsymbol{\beta^{*}}}\left(y\right)g_{\boldsymbol{\beta}}^{\left(\textrm{I}\right)}\left(\boldsymbol{\beta}\right)d\boldsymbol{\beta}
\]
\begin{equation}
={\displaystyle \int_{\boldsymbol{\mathcal{B}}\left(k\right)}}F_{Y\mid\boldsymbol{\beta};\boldsymbol{\beta^{*}}}\left(y\right)g_{\boldsymbol{\beta}}^{\left(\textrm{II}\right)}\left(\boldsymbol{\beta}\right)d\boldsymbol{\beta}
\end{equation}
for all $y\in\mathfrak{Y}$, where $g_{\boldsymbol{\beta}}^{\left(\textrm{I}\right)}\left(\boldsymbol{\beta}\right)\neq g_{\boldsymbol{\beta}}^{\left(\textrm{II}\right)}\left(\boldsymbol{\beta}\right)$
for all $\boldsymbol{\beta}$ within some continuous region $\boldsymbol{\mathcal{B}}\subseteq\boldsymbol{\mathcal{B}}\left(k\right)$
of dimension $k$.

Given $\textrm{M}_{Y\mid\boldsymbol{\beta};\boldsymbol{\beta^{*}}}\left(t\right)$,
we can re-express (2) through the equivalent condition,
\[
\textrm{M}_{Y\mid\boldsymbol{\beta^{*}}}\left(t\right)={\displaystyle \int_{\boldsymbol{\mathcal{B}}\left(k\right)}}\textrm{M}_{Y\mid\boldsymbol{\beta};\boldsymbol{\beta^{*}}}\left(t\right)g_{\boldsymbol{\beta}}^{\left(\textrm{I}\right)}\left(\boldsymbol{\beta}\right)d\boldsymbol{\beta}
\]
\begin{equation}
={\displaystyle \int_{\boldsymbol{\mathcal{B}}\left(k\right)}}\textrm{M}_{Y\mid\boldsymbol{\beta};\boldsymbol{\beta^{*}}}\left(t\right)g_{\boldsymbol{\beta}}^{\left(\textrm{II}\right)}\left(\boldsymbol{\beta}\right)d\boldsymbol{\beta}
\end{equation}
for all $t$ in some interval $\left[0,\epsilon_{2}^{\prime}\right)\subseteq\left[0,\epsilon_{2}\right)$.
Then, substituting $\boldsymbol{\beta}=\boldsymbol{\eta}\left(\boldsymbol{\alpha}\right)$
(from Definition 1(i)) into (3) yields
\[
\textrm{M}_{Y\mid\boldsymbol{\beta^{*}}}\left(t\right)={\displaystyle \int_{\boldsymbol{\mathcal{A}}\left(k\right)}}\textrm{M}_{Y\mid\boldsymbol{\eta}\left(\boldsymbol{\alpha}\right);\boldsymbol{\beta^{*}}}\left(t\right)g_{\boldsymbol{\alpha}}^{\left(\textrm{I}\right)}\left(\boldsymbol{\alpha}\right)d\boldsymbol{\alpha}
\]
\begin{equation}
={\displaystyle \int_{\boldsymbol{\mathcal{A}}\left(k\right)}}\textrm{M}_{Y\mid\boldsymbol{\eta}\left(\boldsymbol{\alpha}\right);\boldsymbol{\beta^{*}}}\left(t\right)g_{\boldsymbol{\alpha}}^{\left(\textrm{II}\right)}\left(\boldsymbol{\alpha}\right)d\boldsymbol{\alpha},
\end{equation}
where
\[
g_{\boldsymbol{\alpha}}^{\left(\textrm{I}\right)}\left(\boldsymbol{\alpha}\right)=g_{\boldsymbol{\beta}}^{\left(\textrm{I}\right)}\left(\boldsymbol{\eta}\left(\boldsymbol{\alpha}\right)\right)\left|\det\left(\mathbf{J}_{\boldsymbol{\eta}}\right)\right|
\]
\[
\neq g_{\boldsymbol{\alpha}}^{\left(\textrm{II}\right)}\left(\boldsymbol{\alpha}\right)=g_{\boldsymbol{\beta}}^{\left(\textrm{II}\right)}\left(\boldsymbol{\eta}\left(\boldsymbol{\alpha}\right)\right)\left|\det\left(\mathbf{J}_{\boldsymbol{\eta}}\right)\right|
\]
(with $\mathbf{J}_{\boldsymbol{\eta}}$ denoting the Jacobian matrix
associated with $\boldsymbol{\eta}$) for all $\boldsymbol{\alpha}$
within some continuous region $\boldsymbol{\mathcal{A}}\subseteq\boldsymbol{\mathcal{A}}\left(k\right)$
of dimension $k$.

Finally, we substitute $t=\xi\left(s\right)$ (from Definition 1(ii))
into (4), implying (from Definition 1(iii)) that
\[
\textrm{M}_{X\mid\boldsymbol{\alpha^{*}}}\left(s\right)={\displaystyle \int_{\boldsymbol{\mathcal{A}}\left(k\right)}}\textrm{M}_{X\mid\boldsymbol{\alpha};\boldsymbol{\alpha^{*}}}\left(s\right)g_{\boldsymbol{\alpha}}^{\left(\textrm{I}\right)}\left(\boldsymbol{\alpha}\right)d\boldsymbol{\alpha}
\]
\[
={\displaystyle \int_{\boldsymbol{\mathcal{A}}\left(k\right)}}\textrm{M}_{X\mid\boldsymbol{\alpha};\boldsymbol{\alpha^{*}}}\left(s\right)g_{\boldsymbol{\alpha}}^{\left(\textrm{II}\right)}\left(\boldsymbol{\alpha}\right)d\boldsymbol{\alpha},
\]
where $g_{\boldsymbol{\alpha}}^{\left(\textrm{I}\right)}\left(\boldsymbol{\alpha}\right)\neq g_{\boldsymbol{\alpha}}^{\left(\textrm{II}\right)}\left(\boldsymbol{\alpha}\right)$
for all $\boldsymbol{\alpha}$ in $\boldsymbol{\mathcal{A}}\subseteq\boldsymbol{\mathcal{A}}\left(k\right)$
and $s\in\left[0,\epsilon_{1}^{\prime}\right)\subseteq\left[0,\epsilon_{1}\right)$.
This shows that if $F_{Y\mid\boldsymbol{\beta};\boldsymbol{\beta^{*}}}$
does not always generate identifiable continuous mixtures, then $F_{X\mid\boldsymbol{\alpha};\boldsymbol{\alpha^{*}}}$
cannot do so either.

Since GF accessibility is symmetric with respect to $\textrm{M}_{X\mid\boldsymbol{\alpha};\boldsymbol{\alpha^{*}}}\left(s\right)$
and $\textrm{M}_{Y\mid\boldsymbol{\beta};\boldsymbol{\beta^{*}}}\left(t\right)$,
the ``only if'' part of the theorem (i.e., that if $F_{Y\mid\boldsymbol{\beta};\boldsymbol{\beta^{*}}}$
generates identifiable continuous mixtures, then $F_{X\mid\boldsymbol{\alpha};\boldsymbol{\alpha^{*}}}$
must do so as well) follows by an analogous argument. $\blacksquare$\medskip{}

\subsection{Special Case: Poisson$\boldsymbol{\left(\lambda\right)}$ Kernel}

\noindent The $\textrm{Poisson}\left(\lambda\right)$ kernel with
free $\lambda$ provides a useful starting point for applying Theorem
1 because its associated MGF consists of a simple double-exponential
function, $\textrm{M}_{X\mid\lambda}^{\left(\textrm{P}\right)}\left(s\right)=\exp\left(\lambda\left(e^{s}-1\right)\right)$,
from which many other MGFs are GF accessible. This observation is
formalized in the following corollary to Theorem 1.

\medskip{}

\noindent \textbf{Corollary 1:} If $Y\mid\beta;\boldsymbol{\beta^{*}}\sim F_{Y\mid\beta;\boldsymbol{\beta^{*}}}\left(y\right)$
with free parameter $\beta$, fixed parameter vector $\boldsymbol{\beta^{*}}$,
and
\begin{equation}
\textrm{M}_{Y\mid\beta;\boldsymbol{\beta^{*}}}\left(t\right)=\exp\left(\eta^{-1}\left(\beta\right)\xi^{-1}\left(t\right)\right)
\end{equation}
for continuous one-to-one mappings $\eta$ and $\xi$, then all continuous
mixtures generated by the kernel $F_{Y\mid\beta;\boldsymbol{\beta^{*}}}$
are identifiable.\medskip{}

\noindent \textbf{Proof:} Given that $\textrm{M}_{X\mid\lambda}^{\left(\textrm{P}\right)}\left(s\right)=\exp\left(\lambda\left(e^{s}-1\right)\right)$
for $X\mid\lambda\sim\textrm{Poisson}\left(\lambda\right)$, \foreignlanguage{american}{we
can rewrite (5) as}
\begin{equation}
\textrm{M}_{Y\mid\beta;\boldsymbol{\beta^{*}}}\left(t\right)=\exp\left(\eta^{-1}\left(\beta\right)\left[\exp\left(\xi^{\prime-1}\left(t\right)\right)-1\right]\right),
\end{equation}
where $\lambda=\eta^{-1}\left(\beta\right)$ and $s=\xi^{\prime-1}\left(t\right)=\ln\left(\xi^{-1}\left(t\right)+1\right)$\foreignlanguage{american}{.
Since (6) in turn is equivalent to condition (iii) of Definition 1,
it is clear that $\textrm{M}_{Y\mid\beta;\boldsymbol{\beta^{*}}}\left(t\right)$
is GF accessible from $\textrm{M}_{X\mid\lambda}^{\left(\textrm{P}\right)}\left(s\right)$,
and the desired result follows from Theorem 1.} $\blacksquare$\medskip{}

It is important to note that (5) also can be written as
\[
\textrm{M}_{Y\mid\beta;\boldsymbol{\beta^{*}}}\left(t\right)=\left[\exp\left(\xi^{-1}\left(t\right)\right)\right]^{\eta^{-1}\left(\beta\right)}.
\]
Thus, Corollary 1 may be used to demonstrate the identifiability of
continuous mixtures of kernels from the additively closed family (see
Subsection 2.2) for which the MGF is well defined. 

\subsection{Special Case: Scale-Parameter Kernels}

\noindent The scale-parameter family provides another fertile starting
point for generating results from Theorem 1 because any kernel $F_{X\mid\theta}^{\left(\textrm{SP}\right)}$
belonging to this family (with free $\theta$) is characterized by
an MGF with the simple form $\textrm{M}_{X\mid\theta}^{\left(\textrm{SP}\right)}\left(s\right)=\psi_{X\mid\theta}\left(\theta s\right)$
for some continuous function $\psi_{X\mid\theta}$ defined on $\mathbb{R}^{+}$.\footnote{The indicated functional form may be derived as follows. First, let
$\textrm{M}_{X\mid\theta}^{\left(\textrm{SP}\right)}\left(s\right)=\textrm{E}_{X\mid\theta}\left[e^{sX}\right]=h\left(\theta,s\right)$
for some continuous bivariate function $h$. Then, since $F_{X\mid\theta}^{\left(\textrm{SP}\right)}$
belongs to the scale-parameter family, we know $\textrm{M}_{X\mid\theta}^{\left(\textrm{SP}\right)}\left(\tfrac{s}{\theta}\right)=h\left(1,s\right)$.
Now note that $\textrm{M}_{X\mid\theta}^{\left(\textrm{SP}\right)}\left(s\right)=\textrm{M}_{X\mid\theta}^{\left(\textrm{SP}\right)}\left(\tfrac{\left(\theta s\right)}{\theta}\right)=h\left(1,\theta s\right)$,
which depends on $\theta$ and $s$ through only their product. In
other words, $\psi_{X\mid\theta}\left(\theta s\right)=h\left(1,\theta s\right)$.} This observation yields a second corollary to Theorem 1.

\medskip{}

\noindent \textbf{Corollary 2:} Let $\textrm{M}_{X\mid\theta;\boldsymbol{\alpha^{*}}}^{\left(\textrm{SP}\right)}\left(s\right)=\psi_{X\mid\theta;\boldsymbol{\alpha^{*}}}\left(\theta s\right)$
denote the MGF of a member of the scale-parameter family with free
parameter $\theta$ and fixed parameter vector $\boldsymbol{\alpha^{*}}$.
If $Y\mid\beta;\boldsymbol{\beta^{*}}\sim F_{Y\mid\beta;\boldsymbol{\beta^{*}}}\left(y\right)$
with free parameter $\beta$, fixed parameter vector $\boldsymbol{\beta^{*}}$,
and
\begin{equation}
\textrm{M}_{Y\mid\beta;\boldsymbol{\beta^{*}}}\left(t\right)=\psi_{X\mid\theta;\boldsymbol{\alpha^{*}}}\left(\eta^{-1}\left(\beta\right)\xi^{-1}\left(t\right)\right)
\end{equation}
for continuous one-to-one mappings $\eta$ and $\xi$, then all continuous
mixtures generated by the kernel $F_{Y\mid\beta;\boldsymbol{\beta^{*}}}$
are identifiable.\medskip{}

\noindent \textbf{Proof:} Given that $\textrm{M}_{X\mid\theta;\boldsymbol{\alpha^{*}}}^{\left(\textrm{SP}\right)}\left(s\right)=\psi_{X\mid\theta;\boldsymbol{\alpha^{*}}}\left(\theta s\right)$,
we can set $\theta=\eta^{-1}\left(\beta\right)$ and $s=\xi^{-1}\left(t\right)$
to make\foreignlanguage{american}{ (7) equivalent to condition (iii)
of Definition 1. This implies $\textrm{M}_{Y\mid\beta;\boldsymbol{\beta^{*}}}\left(t\right)$
is GF accessible from $\textrm{M}_{X\mid\theta;\boldsymbol{\alpha^{*}}}^{\left(\textrm{SP}\right)}\left(s\right)$,
and the desired result follows from Theorem 1.} $\blacksquare$\medskip{}

Despite its similarity to Corollary 1, Corollary 2 is not a true generalization
of the earlier result because it fails to include $\psi_{X\mid\theta;\boldsymbol{\alpha^{*}}}\left(\theta s\right)=\exp\left(\theta s\right)$
as a special case (since $\textrm{M}_{X\mid\theta;\boldsymbol{\alpha^{*}}}\left(s\right)=\exp\left(\theta s\right)$
is the MGF of the degenerate distribution with a single point mass
at $X=\theta$, which does not belong to the scale-parameter family).
One particularly effective application of Corollary 2 is to furnish
identifiability results for analogues of scale-parameter distributions
defined on sample spaces other than $\mathbb{R}^{+}$ (e.g., $\mathbb{\mathbb{Z}}$,
$\mathbb{R},$ and $\left(0,1\right)$),\footnote{Such analogues are not unique. See Alzaatreh, Lee, and Famoye (2012),
Buddana and Kozubowski (2014), and Chakraborty (2015) for various
methods of deriving discrete analogues (on subsets of $\mathbb{Z}$),
and Mazucheli et al. (2020) for observations regarding continuous
analogues on $\left(0,1\right)$.} as shown in Subsections 4.2 and 4.3 below. However, it is important
to note that Theorem 1 may be applied directly to achieve results
for analogues of non-scale-parameter distributions, as shown in Subsection
4.4.

\section{Examples}

\noindent To illustrate the usefulness of Theorem 1, we now present
applications based on the following kernel distributions: Poisson;
Gamma; Exponential; Laplace; and Uniform. In the first two cases (Poisson
and Gamma), the results are already known to the published literature,
but helpful in clarifying relationships between the proposed approach
and previously developed methods.

\subsection{Poisson$\boldsymbol{\left(\lambda\right)}$ $\boldsymbol{\rightarrow}$
Normal$\boldsymbol{\left(m,\kappa m\right)}$}

\noindent Let $X\mid\lambda\sim\textrm{Poisson}\left(\lambda\right)$
with free parameter $\lambda$ and associated MGF
\[
\textrm{M}_{X\mid\lambda}^{\left(\textrm{P}\right)}\left(s\right)=\exp\left(\lambda\left(e^{s}-1\right)\right).
\]
Setting $m=\beta\left(\lambda\right)=\lambda$ and $t=\xi\left(s\right)$
such that $s=\xi^{-1}\left(t\right)=\ln\left(\dfrac{1}{2}\kappa t^{2}+t+1\right)$,
for fixed constant $\kappa\in\mathbb{R}^{+}$, then yields $Y\mid m;\kappa\sim\textrm{Normal}\left(m,\varsigma^{2}\right)$
with $\varsigma^{2}=\kappa m$ and MGF
\begin{equation}
\textrm{M}_{Y\mid m;\kappa}^{\left(\textrm{N}\right)}\left(t\right)=\exp\left(m\left(t+\dfrac{1}{2}\kappa t^{2}\right)\right),
\end{equation}
and the identifiability of $\textrm{Normal}\left(m,\kappa m\right)$
continuous mixtures follows from Corollary 1. Since (8) can be re-expressed
as
\[
\textrm{M}_{Y\mid m;\kappa}^{\left(\textrm{N}\right)}\left(t\right)=\left[\exp\left(t+\dfrac{1}{2}\kappa t^{2}\right)\right]^{m},
\]
the same result may be demonstrated by noting that $\textrm{Normal}\left(m,\kappa m\right)$
belongs to the additively closed family (consistent with prior comments
in Subsection 3.3).

The $\textrm{Normal}\left(m,\kappa m\right)$ kernel forms the basis
for the ``Normal mean-variance mixture'' model, which appears in
geoscientific research of various types. A notable early application
was made by Barndorff-Nielsen (1977), who studied the distribution
of particle sizes of desert sand. Subsequent developments in both
methodology and area of application were addressed by Barndorff-Nielsen,
Kent, and Sørensen (1982). Recently, Louarn, et al. (2024) employed
this family of models to analyze the distribution of non-thermal characteristics
of solar-wind protons.

\subsection{Gamma$\boldsymbol{\left(r,\theta\right)}$ $\boldsymbol{\rightarrow}$
Negative Binomial$\boldsymbol{\left(r,p\right)}$}

\noindent For $X\mid r,\theta\sim\textrm{Gamma}\left(r,\theta\right)$
with associated MGF
\[
\textrm{M}_{X\mid r,\theta}^{\left(\Gamma\right)}\left(s\right)=\left(\dfrac{1}{1-\theta s}\right)^{r},
\]
the most salient discrete analogue is $Y\mid r,p\sim\textrm{Negative Binomial}\left(r,p\right)$
with MGF
\[
\textrm{M}_{Y\mid r,p}^{\left(\textrm{NB}\right)}\left(t\right)=\left(\dfrac{1-p}{1-pe^{t}}\right)^{r}.
\]
Setting $p=\eta\left(\theta\right)=\tfrac{\theta}{1+\theta}$, $t=\xi\left(s\right)=\ln\left(s+1\right)$,\footnote{As noted by Buddana and Kozubowski (2014), the transformation $t=\xi\left(s\right)=\ln\left(s+1\right)$
is commonly used to derive discrete analogues from continuous distributions. } and $\psi_{X\mid r,\theta}\left(\theta s\right)=\left(\tfrac{1}{1-\theta s}\right)^{r}$
in Corollary 2 then confirms the identifiability of $\textrm{Negative Binomial}\left(r,p\right)$
continuous mixtures with free $p$ and fixed $r$ (which is already
known from Subsection 2.5).

The $\textrm{Negative Binomial}\left(r,p\right)$ distribution is
routinely deployed to model event counts in risk analysis, especially
in the context of extreme weather (see, e.g., Vitolo and Stephenson,
2009). Often, it is preferred to the $\textrm{Poisson}\left(\lambda\right)$
distribution, both because of the former's overdispersion property
(i.e., $\textrm{Var}_{X\mid p;r}^{\left(\textrm{NB}\right)}\left[X\right]>\textrm{E}_{X\mid p;r}^{\left(\textrm{NB}\right)}\left[X\right]$
for all parameter values) and additional flexibility from possessing
a second (fixed) parameter. In Powers and Xu (2025), the present authors
employ $\textrm{Negative Binomial}\left(r,p\right)$ as the ``canonical''
kernel for event-count mixtures, recognizing that $\textrm{Poisson}\left(\lambda\right)$
itself can be formed as a continuous Negative Binomial mixture.

\subsection{Exponential$\boldsymbol{\left(\theta\right)}$ $\boldsymbol{\rightarrow}$
Laplace$\boldsymbol{\left(0,\varsigma\right)}$}

\noindent Consider $X\mid\theta\sim\textrm{Exponential}\left(\theta\right)$
with free $\theta$, a special case of $\textrm{Gamma}\left(r,\theta\right)$
(for $r=1$) with MGF
\[
\textrm{M}_{X\mid\theta}^{\left(\textrm{E}\right)}\left(s\right)=\dfrac{1}{1-\theta s}.
\]
Furthermore, let $Y=\left(2B-1\right)X$, where $B\sim\textrm{Bernoulli}\left(\tfrac{1}{2}\right)$
is statistically independent of $X$, yielding $Y\mid\varsigma\sim\textrm{Laplace}\left(m,\varsigma\right)$
with free $\varsigma$ and $m=0$ as an analogue of $\textrm{Exponential}\left(\theta\right)$
(defined on $\mathbb{R}$) with MGF
\begin{equation}
\textrm{M}_{Y\mid\varsigma}^{\left(\textrm{L}\right)}\left(t\right)=\dfrac{1}{1-\varsigma^{2}t^{2}}.
\end{equation}
Setting $\varsigma=\eta\left(\theta\right)=\sqrt{\theta}$, $t=\xi\left(s\right)=\sqrt{s}$,
and $\psi_{X\mid\theta}\left(\theta s\right)=\tfrac{1}{1-\theta s}$
in Corollary 2 then shows that continuous mixtures of $F_{Y\mid\varsigma}^{\left(\textrm{L}\right)}$
are identifiable.

The $\textrm{Laplace}\left(0,\varsigma\right)$ distribution is commonly
used to model incremental changes in the magnitudes of geophysical
processes. This is intuitively reasonable because the intensities
of many naturally occurring processes are themselves $\textrm{Exponential}\left(\theta\right)\equiv\textrm{Gamma}\left(1,\theta\right)$,
and the difference between two independent $\textrm{Exponential}\left(\theta\right)$
random variables is $\textrm{Laplace}\left(0,\theta\right)$. Recent
applications of the latter distribution include van der Elst (2021),
for successive earthquake magnitudes, and Ito, Ohba, and Mizuno (2024),
for changes in airplane headwind magnitudes.

\subsection{Laplace$\boldsymbol{\left(0,\varsigma\right)}$ $\boldsymbol{\rightarrow}$
Discrete Laplace$\boldsymbol{\left(p\right)}$}

\noindent From Inusah and Kozubowski (2006), it is known that $X\mid\varsigma\sim\textrm{Laplace}\left(0,\varsigma\right)$
possesses a discrete analogue on $\mathbb{Z}$: $Y\mid p\sim\textrm{Discrete Laplace}\left(p\right)$
with free $p\in\left(0,1\right)$. This model is characterized by
the following CDF and MGF, respectively:
\[
F_{Y\mid p}^{\left(\textrm{DL}\right)}\left(y\right)=\begin{cases}
\dfrac{p^{-y}}{1+p} & \textrm{for }y\in\mathbb{Z}^{-}\\
1-\dfrac{p^{y+1}}{1+p} & \textrm{for }y\in\mathbb{Z}_{0}^{+}
\end{cases};\textrm{ and}
\]
\[
\textrm{M}_{Y\mid p}^{\left(\textrm{DL}\right)}\left(t\right)=\dfrac{\left(1-p\right)^{2}}{\left(1-pe^{t}\right)\left(1-pe^{-t}\right)}.
\]
Rewriting (9) as
\[
\textrm{M}_{X\mid\varsigma}^{\left(\textrm{L}\right)}\left(s\right)=\dfrac{1}{1-\varsigma^{2}s^{2}}
\]
and setting $p=\eta\left(\varsigma\right)=1+\tfrac{1-\sqrt{2\varsigma^{2}+1}}{\varsigma^{2}}$
and $t=\xi\left(s\right)=\cosh^{-1}\left(s^{2}+1\right)$ then allows
us to conclude from Theorem 1 that continuous mixtures of $F_{Y\mid p}^{\left(\textrm{D}\right)}$
are identifiable.

As noted in the prior subsection, the $\textrm{Laplace}\left(0,\varsigma\right)$
distribution may be used to model incremental changes in geophysical-process
magnitudes. Since such intensities often are rounded to discrete numerical
values, it is increasingly common to use the $\textrm{Discrete Laplace}\left(p\right)$
distribution instead. See van der Elst (2021) and Tinti and Gasperini
(2024) for discussions of how this distribution is implemented in
practice.

\subsection{Uniform$\boldsymbol{\left(a,b\right)}$ $\boldsymbol{\rightarrow}$
Differentiated Error Function$\boldsymbol{\left(a,b\right)}$}

\noindent We now consider the case of $X\mid a,b\sim\textrm{Uniform}\left(a,b\right)\Longleftrightarrow F_{X\mid a,b}^{\left(\textrm{U}\right)}\left(x\right)=\tfrac{x-a}{b-a},\:x\in\left(a,b\right)$
with free $\left(a,b\right)\in\left(\mathbb{R}^{+}\right)^{2}$, for
which the MGF is
\[
\textrm{M}_{X\mid a,b}^{\left(\textrm{U}\right)}\left(s\right)=\dfrac{e^{bs}-e^{as}}{\left(b-a\right)s}.
\]
This example is of interest because continuous mixtures of the $\textrm{Uniform}\left(a,b\right)$
distribution are identifiable if exactly one of the parameters $m=\tfrac{a+b}{2}$
(the midpoint of the sample space) or $\ell=b-a$ (the length of the
sample space) is free, but not if both are free (see Teicher, 1961).
Naturally, the unidentifiability of continuous mixtures when both
$m$ and $\ell$ are free implies the same result for free $\left(a,b\right)$.

Setting $\boldsymbol{\eta}\left(a,b\right)=\left(a,b\right)$ and
$t=\xi\left(s\right)=\sqrt{s}$ yields
\begin{equation}
{\displaystyle \textrm{M}_{Y\mid a,b}^{\left(\textrm{DE}\right)}\left(t\right)=\dfrac{\exp\left(bt^{2}\right)-\exp\left(at^{2}\right)}{\left(b-a\right)t^{2}}}
\end{equation}
for $Y\mid a,b\sim F_{Y\mid a,b}^{\left(\textrm{DE}\right)}\left(y\right),\:y\in\mathbb{R}$
with free $\left(a,b\right)$, where we will say $Y\mid a,b$ possesses
a $\textrm{Differentiated Error Function}\left(a,b\right)$ distribution
(because of the functional form of its PDF, shown in (13) below).
The unidentifiability of continuous mixtures of $F_{Y\mid a,b}^{\left(\textrm{DE}\right)}$
then follows from Theorem 1.

Since this new distribution is not well known, we solve for its PDF
via CF inversion.\footnote{The PDF is used instead of the CDF because of greater analytical tractability.}
To this end, consider the CF counterpart of (10),
\begin{equation}
{\displaystyle \varphi_{Y\mid a,b}^{\left(\textrm{DE}\right)}\left(\omega\right)=\dfrac{\exp\left(-a\omega^{2}\right)-\exp\left(-b\omega^{2}\right)}{\left(b-a\right)\omega^{2}}},
\end{equation}
and apply the approach of Gil-Pelaez (1951) to obtain
\[
f_{Y\mid a,b}^{\left(\textrm{DE}\right)}\left(y\right)=\dfrac{1}{2\pi}{\displaystyle \int_{0}^{\infty}\left[e^{iy\omega}\varphi_{Y\mid a,b}^{\left(\textrm{DE}\right)}\left(-\omega\right)+e^{-iy\omega}\varphi_{Y\mid a,b}^{\left(\textrm{DE}\right)}\left(\omega\right)\right]d\omega}
\]
\begin{equation}
=\dfrac{1}{\pi\left(b-a\right)}{\displaystyle \int_{0}^{\infty}\dfrac{\cos\left(y\omega\right)\left[\exp\left(-a\omega^{2}\right)-\exp\left(-b\omega^{2}\right)\right]}{\omega^{2}}d\omega}.
\end{equation}
As shown in the Appendix, this simplifies to
\[
f_{Y\mid a,b}^{\left(\textrm{DE}\right)}\left(y\right)=\dfrac{y}{2\left(b-a\right)}\left[\textrm{erf}\left(\dfrac{y}{2\sqrt{b}}\right)-\textrm{erf}\left(\dfrac{y}{2\sqrt{a}}\right)\right]
\]
\begin{equation}
+\dfrac{1}{2\left(b-a\right)}\left[\sqrt{b}\:\textrm{erf}\,^{\prime}\left(\dfrac{y}{2\sqrt{b}}\right)-\sqrt{a}\:\textrm{erf}\,^{\prime}\left(\dfrac{y}{2\sqrt{a}}\right)\right],
\end{equation}
where $\textrm{erf}\left(u\right)=\tfrac{2}{\sqrt{\pi}}{\textstyle \int_{0}^{u}}\exp\left(-z^{2}\right)dz$
denotes the error function.

Unlike the $\textrm{Uniform}\left(a,b\right)$ distribution, whose
sample space, $\left(a,b\right)$, is both bounded and dependent on
its parameter values, the $\textrm{Differentiated Error Function}\left(a,b\right)$
distribution has sample space $\mathfrak{Y}=\mathbb{R}$, which is
unbounded and therefore independent of $a$ and $b$. Moreover, the
MGF in (10) yields
\[
{\displaystyle \textrm{E}_{Y\mid a,b}^{\left(\textrm{DE}\right)}\left[Y\right]=0}
\]
and
\[
{\displaystyle \textrm{Var}_{Y\mid a,b}^{\left(\textrm{DE}\right)}\left[Y\right]}=a+b,
\]
which are quite different from $\textrm{E}_{X\mid a,b}^{\left(\textrm{U}\right)}\left[X\right]$
and $\textrm{Var}_{X\mid a,b}^{\left(\textrm{U}\right)}\left[X\right]$,
respectively. Consequently, $F_{Y\mid a,b}^{\left(\textrm{DE}\right)}$
does not appear to be an ``analogue'' of $F_{X\mid a,b}^{\left(\textrm{U}\right)}$
in any practical sense.

Figure 1 below presents plots of $f_{Y\mid a,b}^{\left(\textrm{DE}\right)}\left(y\right)$
for various values of the distribution's parameters. These figures
reveal that, for all $a>0$, the PDF possesses a smooth, unimodal
shape symmetric about 0. In the limiting case of $a=0$, the PDF manifests
a sharp peak at the origin, where it is not continuously differentiable.
Furthermore, the plots are consistent with the observation that the
variance of the distribution increases over $a+b$, and is sensitive
to the length of the corresponding $\textrm{Uniform}\left(a,b\right)$
sample space, $\ell=b-a$, only indirectly (i.e., for fixed $a$,
greater values of $b-a$ are associated with greater values of $a+b$).
\noindent \begin{center}
\includegraphics[scale=0.14]{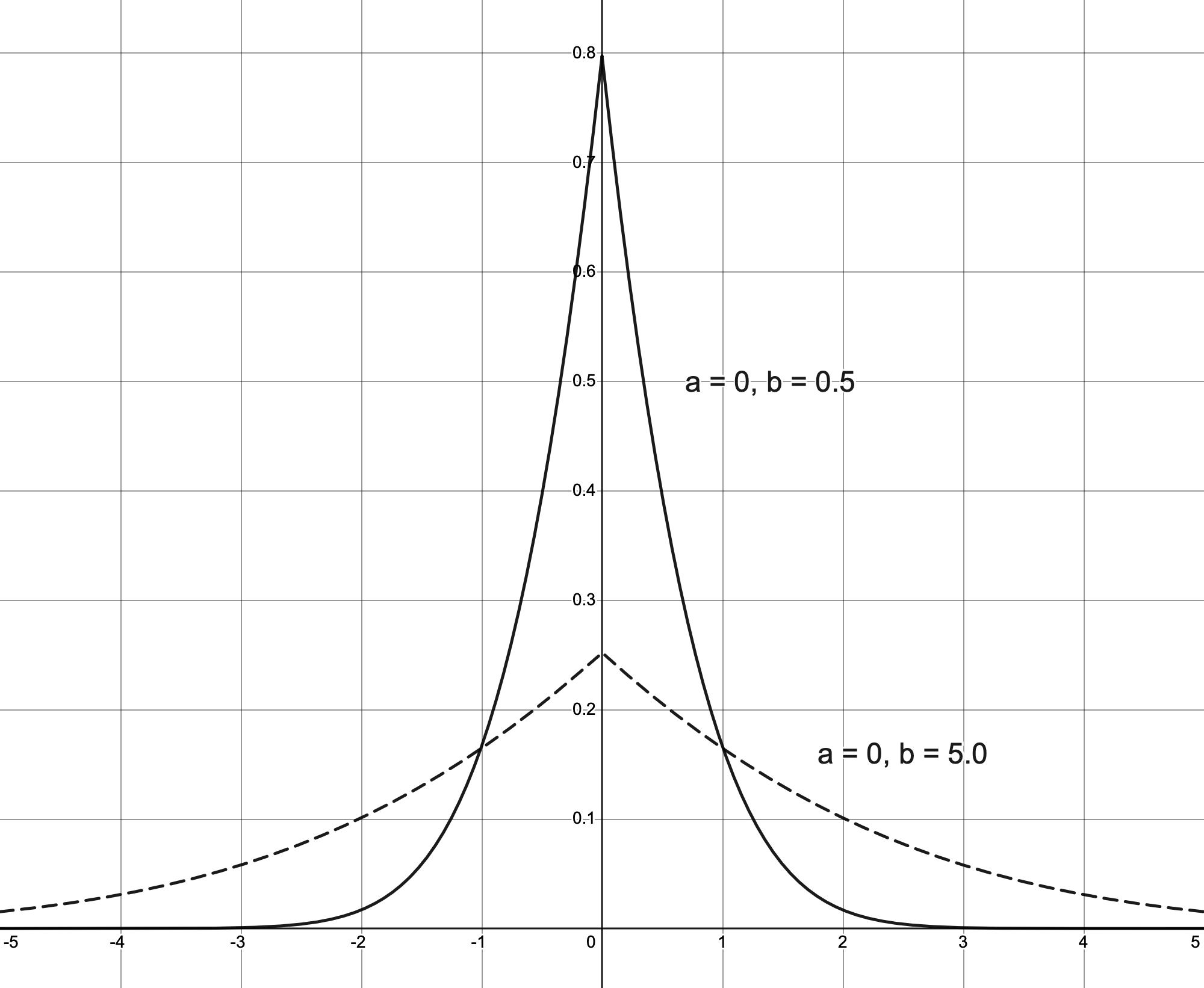}
\par\end{center}

\noindent \begin{center}
Figure 1a. $\textrm{Differentiated Error Function}\left(a,b\right)$
PDFs\medskip{}
\par\end{center}

\noindent \begin{center}
\includegraphics[scale=0.14]{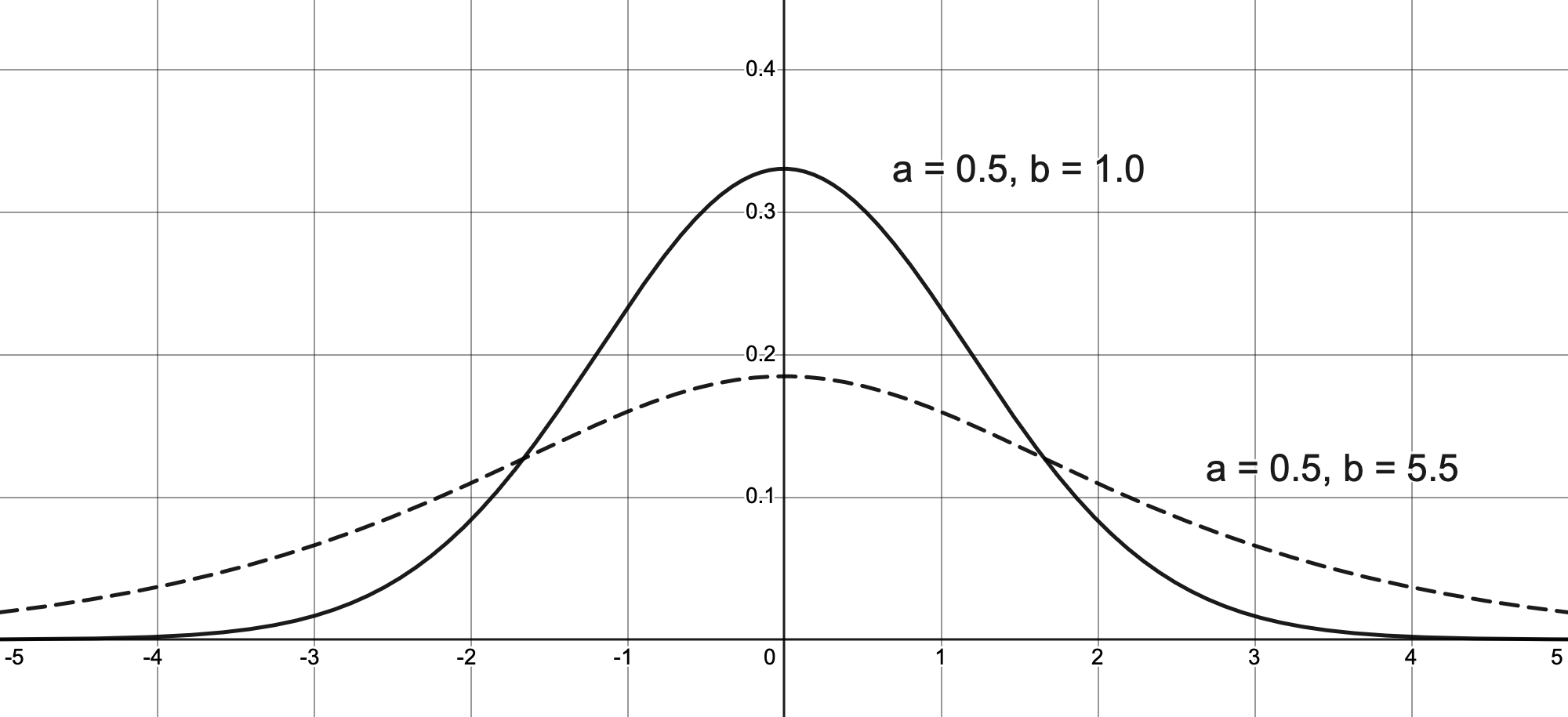}
\par\end{center}

\noindent \begin{center}
Figure 1b. $\textrm{Differentiated Error Function}\left(a,b\right)$
PDFs\medskip{}
\newpage{}
\par\end{center}

\noindent \begin{center}
\includegraphics[scale=0.14]{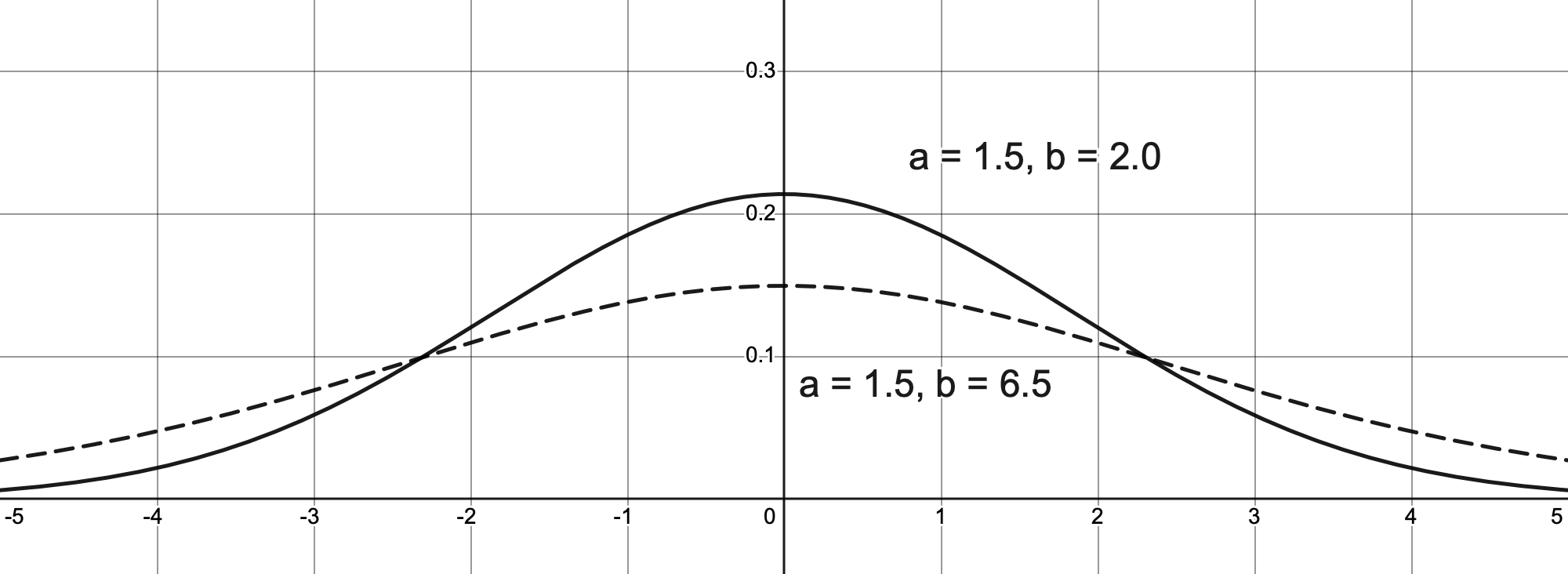}
\par\end{center}

\noindent \begin{center}
Figure 1c. $\textrm{Differentiated Error Function}\left(a,b\right)$
PDFs
\par\end{center}

From the above plots and the PDF in (13), it is clear that the $\textrm{Differentiated Error Function}\left(a,b\right)$
distribution can serve as a viable alternative to either $\textrm{Laplace}\left(0,\varsigma\right)$
or $\textrm{Normal}\left(0,\varsigma^{2}\right)$, depending on the
circumstances. To see this, first consider that $f_{Y\mid a,b}^{\left(\textrm{DE}\right)}\left(y\right)=O\left(y^{-2}\exp\left(-\tfrac{y^{2}}{4b}\right)\right)$
as $\left|y\right|\rightarrow\infty$, whereas $f_{Y\mid0;\varsigma^{2}}^{\left(\textrm{N}\right)}\left(y\right)=O\left(\exp\left(-\tfrac{y^{2}}{2\varsigma^{2}}\right)\right)$
and $f_{Y\mid0;\varsigma}^{\left(\textrm{L}\right)}\left(y\right)=O\left(\exp\left(-\tfrac{\left|y\right|}{\varsigma}\right)\right)$.
It then follows that: (i) if observed data possess a Laplace-like
peak near 0, but substantially lighter-than-Laplace tails, then $\textrm{Differentiated Error Function}\left(a,b\right)$
may be appropriate; and (ii) if data are Normal-like except for slightly
lighter tails, then $\textrm{Differentiated Error Function}\left(a,b\right)$
again may offer a good fit.

\section{Conclusion}

\noindent Mixture distributions are widely employed in the natural
and social sciences, and particularly well suited for modeling complex
geoscientific processes and their attendant risks. In such applications,
it usually is desirable \textendash{} and often crucial \textendash{}
to be able to associate any given mixture distribution with a unique
mixing distribution (and vice versa). This identifiability property
has been investigated closely since the early work of Feller (1943)
and Teicher (1960, 1961, and 1963), but broadly applicable results
are somewhat limited.

In the present article, we provided a simple criterion \textendash{}
GF accessibility \textendash{} for extending previously known results
to new kernel distributions. We then illustrated the proposed approach
through results for several specific types of continuous mixtures.
These examples revealed that, although GF accessibility does not offer
explicit sufficient conditions for identifiability (such as those
summarized in Subsections 2.1 through 2.5), it does make two significant
contributions of a general nature.

First, the technique can be used to provide simple alternative proofs
of certain well-known results, as for members of the additively closed
family with well-defined MGFs (Subsections 3.3 and 4.1) and $\textrm{Negative Binomial}\left(r,p\right)$
kernels with free $p$ and fixed $r$ (Subsection 4.2). Second, it
affords the potential for generating new identifiability results;
in particular, for various analogues (either discrete or continuous)
of the original kernel distribution (Subsections 4.2 through 4.4).
Given the non-uniqueness of such analogues, we believe further investigation
of the functional relationships between the MGFs of certain families
and their analogues will facilitate improved methods for understanding
and classifying the latter distributions.

\noindent \newpage{}

\section*{Appendix}

\noindent This appendix provides the derivation of the PDF associated
with the $\textrm{Differentiated Error Function}\left(\theta\right)$
distribution.

First, note that (12) may be expressed as
\[
f_{Y\mid a,b}^{\left(\textrm{DE}\right)}\left(y\right)=\dfrac{A-B}{\pi\left(b-a\right)},\qquad\qquad\quad\textrm{(B1)}
\]
where
\[
A={\displaystyle \int_{0}^{\infty}\dfrac{\cos\left(y\omega\right)\exp\left(-a\omega^{2}\right)}{\omega^{2}}d\omega}\qquad\quad\textrm{(B2)}
\]
and
\[
B={\displaystyle \int_{0}^{\infty}\dfrac{\cos\left(y\omega\right)\exp\left(-b\omega^{2}\right)}{\omega^{2}}d\omega}.\qquad\quad\textrm{(B3)}
\]
Rewriting (B2) and (B3) as
\[
A=\underset{\gamma\rightarrow0^{+}}{\lim}{\displaystyle \int_{0}^{\infty}\dfrac{\cos\left(y\omega\right)\exp\left(-a\omega^{2}\right)}{\omega^{2}+\gamma^{2}}d\omega}
\]
and
\[
B=\underset{\gamma\rightarrow0^{+}}{\lim}{\displaystyle \int_{0}^{\infty}\dfrac{\cos\left(y\omega\right)\exp\left(-b\omega^{2}\right)}{\omega^{2}+\gamma^{2}}d\omega},
\]
respectively, we find from equation 3.954(2) of Gradshteyn and Ryzhik
(2007) that
\[
A-B=\underset{\gamma\rightarrow0^{+}}{\lim}\dfrac{\pi}{4\gamma}\left\{ e^{a\gamma^{2}}\left[2\cosh\left(\gamma y\right)-e^{-\gamma y}\textrm{erf}\left(\gamma\sqrt{a}-\dfrac{y}{2\sqrt{a}}\right)-e^{\gamma y}\textrm{erf}\left(\gamma\sqrt{a}+\dfrac{y}{2\sqrt{a}}\right)\right]\right.
\]
\[
\left.-e^{b\gamma^{2}}\left[2\cosh\left(\gamma y\right)-e^{-\gamma y}\textrm{erf}\left(\gamma\sqrt{b}-\dfrac{y}{2\sqrt{b}}\right)-e^{\gamma y}\textrm{erf}\left(\gamma\sqrt{b}+\dfrac{y}{2\sqrt{b}}\right)\right]\right\} 
\]
\[
=\underset{\gamma\rightarrow0^{+}}{\lim}\dfrac{\pi}{4\gamma}\left\{ e^{a\gamma^{2}}\left[e^{-\gamma y}\left(1-\textrm{erf}\left(\gamma\sqrt{a}-\dfrac{y}{2\sqrt{a}}\right)\right)+e^{\gamma y}\left(1-\textrm{erf}\left(\gamma\sqrt{a}+\dfrac{y}{2\sqrt{a}}\right)\right)\right]\right.
\]
\[
\left.-e^{b\gamma^{2}}\left[e^{-\gamma y}\left(1-\textrm{erf}\left(\gamma\sqrt{b}-\dfrac{y}{2\sqrt{b}}\right)\right)+e^{\gamma y}\left(1-\textrm{erf}\left(\gamma\sqrt{b}+\dfrac{y}{2\sqrt{b}}\right)\right)\right]\right\} 
\]
\[
=\underset{\gamma\rightarrow0^{+}}{\lim}\dfrac{\pi}{4\gamma}\left[e^{a\gamma^{2}-\gamma y}\left(1-\textrm{erf}\left(\gamma\sqrt{a}-\dfrac{y}{2\sqrt{a}}\right)\right)+e^{a\gamma^{2}+\gamma y}\left(1-\textrm{erf}\left(\gamma\sqrt{a}+\dfrac{y}{2\sqrt{a}}\right)\right)\right.
\]
\[
\left.-e^{b\gamma^{2}-\gamma y}\left(1-\textrm{erf}\left(\gamma\sqrt{b}-\dfrac{y}{2\sqrt{b}}\right)\right)-e^{b\gamma^{2}+\gamma y}\left(1-\textrm{erf}\left(\gamma\sqrt{b}+\dfrac{y}{2\sqrt{b}}\right)\right)\right],
\]
where $\textrm{erf}\left(u\right)=\tfrac{2}{\sqrt{\pi}}{\textstyle \int_{0}^{u}}\exp\left(-z^{2}\right)dz$.

Since the limit of the factor in square brackets is 0, it follows
from L'Hôpital's rule that
\[
A-B=\underset{\gamma\rightarrow0^{+}}{\lim}\dfrac{\pi}{4}\left[\left(2a\gamma-y\right)e^{a\gamma^{2}-\gamma y}\left(1-\textrm{erf}\left(\gamma\sqrt{a}-\dfrac{y}{2\sqrt{a}}\right)\right)-\sqrt{a}e^{a\gamma^{2}-\gamma y}\left(\textrm{erf}\,^{\prime}\left(\gamma\sqrt{a}-\dfrac{y}{2\sqrt{a}}\right)\right)\right.
\]
\[
+\left(2a\gamma+y\right)e^{a\gamma^{2}+\gamma y}\left(1-\textrm{erf}\left(\gamma\sqrt{a}+\dfrac{y}{2\sqrt{a}}\right)\right)-\sqrt{a}e^{a\gamma^{2}+\gamma y}\left(\textrm{erf}\,^{\prime}\left(\gamma\sqrt{a}+\dfrac{y}{2\sqrt{a}}\right)\right)
\]
\[
-\left(2b\gamma-y\right)e^{b\gamma^{2}-\gamma y}\left(1-\textrm{erf}\left(\gamma\sqrt{b}-\dfrac{y}{2\sqrt{b}}\right)\right)+\sqrt{b}e^{b\gamma^{2}-\gamma y}\left(\textrm{erf}\,^{\prime}\left(\gamma\sqrt{b}-\dfrac{y}{2\sqrt{b}}\right)\right)
\]
\[
\left.-\left(2b\gamma+y\right)e^{b\gamma^{2}+\gamma y}\left(1-\textrm{erf}\left(\gamma\sqrt{b}+\dfrac{y}{2\sqrt{b}}\right)\right)+\sqrt{b}e^{b\gamma^{2}+\gamma y}\left(\textrm{erf}\,^{\prime}\left(\gamma\sqrt{b}+\dfrac{y}{2\sqrt{b}}\right)\right)\right]
\]
\[
=\dfrac{\pi}{4}\left[-y\left(1-\textrm{erf}\left(-\dfrac{y}{2\sqrt{a}}\right)\right)-\sqrt{a}\left(\textrm{erf}\,^{\prime}\left(-\dfrac{y}{2\sqrt{a}}\right)\right)+y\left(1-\textrm{erf}\left(\dfrac{y}{2\sqrt{a}}\right)\right)-\sqrt{a}\left(\textrm{erf}\,^{\prime}\left(\dfrac{y}{2\sqrt{a}}\right)\right)\right.
\]
\[
\left.+y\left(1-\textrm{erf}\left(-\dfrac{y}{2\sqrt{b}}\right)\right)+\sqrt{b}\left(\textrm{erf}\,^{\prime}\left(-\dfrac{y}{2\sqrt{b}}\right)\right)-y\left(1-\textrm{erf}\left(\dfrac{y}{2\sqrt{b}}\right)\right)+\sqrt{b}\left(\textrm{erf}\,^{\prime}\left(\dfrac{y}{2\sqrt{b}}\right)\right)\right]
\]
\[
=\dfrac{\pi}{4}\left[-2y\:\textrm{erf}\left(\dfrac{y}{2\sqrt{a}}\right)-2\sqrt{a}\left(\textrm{erf}\,^{\prime}\left(\dfrac{y}{2\sqrt{a}}\right)\right)+2y\:\textrm{erf}\left(\dfrac{y}{2\sqrt{b}}\right)+2\sqrt{b}\left(\textrm{erf}\,^{\prime}\left(\dfrac{y}{2\sqrt{b}}\right)\right)\right].
\]
The right-hand side then may be substituted into (B1) to obtain
\[
f_{Y\mid a,b}^{\left(\textrm{DE}\right)}\left(y\right)=\dfrac{1}{2\left(b-a\right)}\left[y\:\textrm{erf}\left(\dfrac{y}{2\sqrt{b}}\right)+\sqrt{b}\:\textrm{erf}\,^{\prime}\left(\dfrac{y}{2\sqrt{b}}\right)-y\:\textrm{erf}\left(\dfrac{y}{2\sqrt{a}}\right)-\sqrt{a}\:\textrm{erf}\,^{\prime}\left(\dfrac{y}{2\sqrt{a}}\right)\right],
\]
which is equivalent to (13).

\begin{thebibliography}{10}
\begin{singlespace}
\bibitem{key-1}Alzaatreh, A., Lee, C., and Famoye, F., 2012, ``On
the Discrete Analogues of Continuous Distributions'', \emph{Statistical
Methodology}, 9, 589-603.

\bibitem{key-21}Barndorff-Nielsen, O., 1977, ``Exponentially Decreasing
Distributions for the Logarithm of Particle Size'', \emph{Proceeding
of the Royal Society of London A}, 353, 1674, 401-419.

\bibitem{key-27}Barndorff-Nielsen, O., Kent, J., and Sørensen, M.,
1982, ``Normal Variance-Mean Mixtures and \emph{z} Distributions'',
\emph{International Statistical Review}, 50, 2, 145-159.

\bibitem{key-3}Bhatti, M. I., 2004, ``On Cluster Effects in Mining
Complex Econometric Data'', in H. Bozdogan, Ed., \emph{Statistical
Data Mining and Knowledge Discovery} (Chapter 23), Chapman and Hall/CRC,
Boca Raton, FL, USA.

\bibitem{key-5}Buddana, A. and Kozubowski, T. J., 2014, ``Discrete
Pareto Distributions'', \emph{Economic Quality Control}, 29, 2, 143-156.

\bibitem{key-3}Chakraborty, S., 2015, ``Generating Discrete Analogues
of Continuous Probability Distributions \textendash{} A Survey of
Methods and Constructions'', \emph{Journal of Statistical Distributions
and Applications}, 2, 6.

\bibitem{key-4}Chaturvedi, A., Bhatti, M. I., Bapat, S. R., and Joshi,
N., 2025, ``Modeling Wind Speed Data Using the Generalized Positive
Exponential Family of Distributions'', \emph{Modeling Earth Systems
and Environment}, 11, 98.

\bibitem{key-11}Cowpertwait, P. S. P. and O\textquoteright Connell,
P. E., 1997, ``A Regionalised Neyman-Scott Model of Rainfall with
Convective and Stratiform Cells'', \emph{Hydrology and Earth System
Sciences}, 1, 71-80.

\bibitem{key-12}Dacre, H. F. and Pinto, J. G., 2020, ``Serial Clustering
of Extratropical Cyclones: A Review of Where, When, and Why It Occurs'',
\emph{npj Climate and Atmospheric Science}, 48.

\bibitem{key-20}Deng, M. and Aminzadeh, M. S., 2023, \textquotedblleft Bayesian
Inference for the Loss models via Mixture Priors'', \emph{Risks},
11, 156. 

\bibitem{key-1}Feller, W., 1943, ``On a General Class of `Contagious'
Distributions'', \emph{Annals of Mathematical Statistics}, 14, 4,
389-400.

\bibitem{key-1}Gil-Pelaez, J., 1951, ``Note on the Inversion Theorem'',
\emph{Biometrika}, 38, 3-4, 481-482.

\bibitem{key-2}Gradshteyn, I. S. and Ryzhik, I. M., 2007, \emph{Tables
of Integrals, Series, and Products}, Seventh Edition, Academic Press,
Burlington, MA, USA.

\bibitem{key-1}Inusah, S. and Kozubowski, T. J., 2006, ``A Discrete
Analogue of the Laplace Distribution'', \emph{Journal of Statistical
Planning and Inference}, 136, 1090-1102.

\bibitem{key-26}Ito, K., Ohba, H., and Mizuno, S., 2024, ``Low-Level
Turbulence Risk Assessment and Visualization Using Temporal Rate of
Change of Headwind of an Aircraft'', \emph{Journal of Big Data},
11, 158.

\bibitem{key-2}Karlis, D. and Xekalaki, E., 2003, ``Mixtures Everywhere'',
in \emph{Stochastic Musings: Perspectives from the Pioneers of the
Late 20th Century}, J. Panaretos, ed., Lawrence Erlbaum Associates,
Mahwah, NJ, USA.

\bibitem{key-19}Li, Y., Tang, N., and Jiang, X., 2016, \textquotedblleft Bayesian
Approaches for Analyzing Earthquake Catastrophic Risk\textquotedblright ,
\emph{Insurance: Mathematics and Economics}, 68, 110-119.

\bibitem{key-22}Louarn, P., Fedorov, A., Prech, L, Owen, C. J., D'Amicis,
R., Bruno, R., \dots , Bale, S. D., 2024, ``Skewness and Kurtosis
of Solar Wind Proton Distribution Functions: The Normal Inverse-Gaussian
Model and Its Implications'', \emph{Astronomy and Astrophysics},
682, A44.

\bibitem{key-2-1}Lüxmann-Ellinghaus, U., 1987, ``On the Identifiability
of Mixtures of Infinitely Divisible Power-Series Distributions'',
\emph{Statistics and Probability Letters}, 5, 375-378.

\bibitem{key-18}Mak, S., Bingham, D., Lu, Y., 2016, ``A Regional
Compound Poisson Process for Hurricane and Tropical Storm Damage'',
\emph{Journal of the Royal Statistical Society, Series C}, 65, 5,
677-703.

\bibitem{key-3}Maritz, J. S and Lwin, T., 1989, \emph{Empirical Bayes
Methods}, Second Edition, CRC Press, Boca Raton, FL, USA.

\bibitem{key-4}Mazucheli, J., Menezes, A. F. B., Fernandes, L. B.,
de Oliveira, R. P., and Ghitany, M. E., 2020, ``The Unit-Weibull
Distribution as an Alternative to the Kumaraswamy Distribution for
the Modeling of Quantiles Conditional on Covariates'', \emph{Journal
of Applied Statistics}, 47, 6, 954-974.

\bibitem{key-8}Neyman, J. and Scott, E., 1952, ``A Theory of the
Spatial Distribution of Galaxies'', \emph{Astrophysical Journal},
116, 144-163.

\bibitem{key-1}Powers, M. R. and Xu, J., 2025, ``Assessing Risk
Heterogeneity through Heavy-Tailed Frequency and Severity Mixtures'',
arXiv:2505.04795.

\bibitem{key-1}Prakasa Rao, B. L. S., 1992, \emph{Identifiability
in Stochastic Models}, Academic Press, Boston, MA, USA.

\bibitem{key-1}Sapatinas, T., 1995, ``Identifiability of Mixtures
of Power-Series Distributions and Related Characterizations'', \emph{Annals
of the Institute of Statistical Mathematics}, 47, 447-459.

\bibitem{key-2}Shlien, S. and Toksöz, M. N., 1970, ``A Clustering
Model for Earthquake Occurrences'', \emph{Bulletin of the Seismological
Society of America}, 60, 6, 1765-1787.

\bibitem{key-2}Stoyanov, J. and Lin, G. W., 2011, ``Mixtures of
Power Series Distributions: Identifiability via Uniqueness in Problems
of Moments'', \emph{Annals of the Institute of Statistical Mathematics},
63, 291-303.

\bibitem{key-15}Tanaka, U., Ogata, Y., and Stoyan, D., 2008, ``Parameter
Estimation and Model Selection for Neyman-Scott Point Processes'',
\emph{Biometrical Journal}, 50, 1, 43-57.

\bibitem{key-1}Teicher, H., 1960, ``On the Mixture of Distributions'',
\emph{Annals of Mathematical Statistics}, 31, 1, 55-73.

\bibitem{key-2}Teicher, H., 1961, ``Identifiability of Mixtures'',
\emph{Annals of Mathematical Statistics}, 32, 1, 244-248.

\bibitem{key-2}Teicher, H., 1963, ``Identifiability of Finite Mixtures'',
\emph{Annals of Mathematical Statistics}, 34, 4, 1265-1269.

\bibitem{key-24}Tinti, S. and Gasperini, P., 2024, ``The Estimation
of \emph{b}-Value of the Frequency-Magnitude Distribution and of Its
1$\sigma$ Intervals from Binned Magnitude Data'', \emph{Geophysical
Journal International}, 238, 433-458.

\bibitem{key-5}van der Elst, N. J., 2021, ``\emph{B-Positive}: A
Robust Estimator of Aftershock Magnitude Distribution in Transiently
Incomplete Catalogs'', \emph{Journal of Geophysical Research: Solid
Earth}, 10.1029/2020JB021027.

\bibitem{key-9}Vere-Jones, D. and Davies, R. B., 1966, ``A Statistical
Survey of Earthquakes in the Main Seismic Region of New Zealand: Part
2 \textendash{} Time Series Analyses'', \emph{New Zealand Journal
of Geology and Geophysics}, 9, 3, 251-284.

\bibitem{key-23}Vitolo, R. and Stephenson, D. B., 2009, ``Serial
Clustering of Intense European Storms'', \emph{Meteorologische Zeitschrift},
18, 411-424.

\bibitem{key-16}Wang, Y., Degleris, A., Williams, A., and Linderman,
S. W., 2024, ``Spatiotemporal Clustering with Neyman-Scott Processes
via Connections to Bayesian Non-Parametric Mixture Models'', \emph{Journal
of the American Statistical Association}, 119, 547, 2382-2395. 

\bibitem{key-6}Younis, F., Aslam, M., and Bhatti, M. I., 2021, ``Preference
of Prior for Two-Component Mixture of Lomax Distribution'', \emph{Journal
of Statistical Theory and Applications}, 20, 2, 407-424.
\end{singlespace}
\end{thebibliography}
\end{document}